\documentclass[conference]{IEEEtran}
\IEEEoverridecommandlockouts
\usepackage{booktabs}
\usepackage{tabularx}
\usepackage[utf8]{inputenc}
\usepackage[T1]{fontenc}
\usepackage{cite}
\usepackage{amsmath,amssymb,amsfonts}
\usepackage{graphicx}
\usepackage{url}
\usepackage{hyperref}
\usepackage{listings}
\usepackage{multirow}
 
\hypersetup{hidelinks}
\begin{document}

\title{Impact of Image Resolution on Age Estimation with DeepFace and InsightFace}

\author{\IEEEauthorblockN{Shiyar Jamo}
\IEEEauthorblockA{Master Applied Artificial Intelligence \\
Amsterdam University of Applied Sciences \\
Amsterdam, Netherlands \\
shiyar.jamo@hva.nl}
}

\maketitle

\begin{abstract}
Automatic age estimation is widely used for age verification, where input images often vary considerably in resolution. This study evaluates the effect of image resolution on age estimation accuracy using DeepFace and InsightFace. A total of 1000 images from the IMDB-Clean dataset were processed in seven resolutions, resulting in 7000 test samples. Performance was evaluated using Mean Absolute Error (MAE), Standard Deviation (SD), and Median Absolute Error (MedAE). This study shows that image resolution has a clear impact on age estimation accuracy. Both frameworks achieve optimal performance at 224x224 pixels, with an MAE of 10.83 years (DeepFace) and 7.46 years (InsightFace). At low resolutions, MAE increases substantially, while very high resolutions also degrade accuracy. InsightFace is consistently faster than DeepFace across all resolutions.
\end{abstract}

  
\section{Introduction}
Automatic age estimation using AI is valuable in many real-world scenarios. 
Examples include self-checkout systems and vending machines that restrict the sale of alcohol or tobacco, online age verification for gaming and social media, and access control at age-restricted events. 
Pretrained frameworks such as DeepFace \cite{serengil_benchmark_2024} and InsightFace \cite{insightface2024} can estimate a person's age from facial images. However, these images may vary considerably in resolution, ranging from low-quality to high-resolution inputs. When images are processed by age estimation frameworks, many systems including DeepFace and InsightFace apply internal preprocessing steps such as resizing or cropping to a fixed input size. 
Common backbone models rely on standard input resolutions, such as 112×112 pixels for ArcFace and 224×224 pixels for VGG-Face \cite{dengArcFace2019, parkhi2015}. These models were originally developed for face recognition \cite{parkhi2015, dengArcFace2019}, but frameworks such as DeepFace repurpose them as feature extractors for related tasks, including age prediction \cite{serengil_benchmark_2024}. Although these preprocessing steps standardize input size, it is unclear whether the resolution of the original image still affects the quality of the final prediction.
However, it remains unclear whether input resolution still influences accuracy despite the internal resizing performed by these frameworks. 
This study uses DeepFace version 0.0.95 and InsightFace version 0.7.3. Previous research has shown that resolution can significantly affect the performance of neural networks in general \cite{kannojia2018}, but no dedicated study has examined how image resolution impacts age estimation specifically within DeepFace and InsightFace. Therefore, this work investigates whether input image resolution continues to affect age estimation accuracy, even though DeepFace and InsightFace internally resize images to fixed dimensions. If input resolution had no influence, all resolutions would yield identical accuracy. If measurable differences occur, this would indicate that external input resolution still affects the final predictions.

\textbf{Research question:}  
What is the impact of input image resolution on the accuracy of age estimation in DeepFace and InsightFace (evaluated using MAE, SD, and MedAE), despite their internal preprocessing pipelines?
\section{Related Work}

\subsection{Background on Age Estimation}
Age estimation from facial images is an important problem within computer vision and pattern recognition. 
The goal is typically to predict the \textit{chronological age} of a person based on a facial photograph. 
A related task is estimating the \textit{apparent age}, which refers to how old a person appears rather than their biological age \cite{zhu2015}.

Researchers have approached this problem using deep neural networks. 
Commonly, large collections of facial images without apparent-age labels are first used to learn a general facial representation, after which the networks are fine-tuned on datasets that do include apparent-age annotations \cite{zhu2015}. 
Nonetheless, the reliability of age estimation remains strongly dependent on the quality and resolution of the input images.

Age estimation can be addressed in two main ways:
\begin{enumerate}
  \item Predicting the \textit{chronological age}, i.e., the actual biological age of the individual;
  \item Predicting the \textit{apparent age}, i.e., how old the person appears in the image \cite{zhu2015, rothe2016}.
\end{enumerate}

In practice, \textit{apparent age} is often used, as it aligns more closely with human perception of age and because these labels are easier to obtain through human annotators \cite{agbo2022}.

\subsection{Regression vs. Classification}
State-of-the-art approaches to age estimation can broadly be divided into two categories:
\begin{enumerate}
    \item \textbf{Regression-based models:}  
    These models treat age as a continuous variable and attempt to predict the exact age (e.g., 24 years). 
    Performance is typically measured using metrics such as Mean Absolute Error (MAE), which captures the average absolute deviation between predicted and true ages \cite{garain2021, chen2019}.
    
    \item \textbf{Classification-based models:}  
    These models treat age as a discrete class or interval (e.g., 20--25 years, 26--30 years). 
    The model predicts the age group rather than an exact value \cite{chen2019}. 
    This approach can be more robust in practice, as individuals of different ages may look very similar. 
    For example, a 28-year-old may visually resemble a 30-year-old; placing both in the same interval (26--30 years) helps reduce ambiguity.
\end{enumerate}

\subsection{Age Estimation Processing Pipeline}

\begin{figure}[h]
  \centering
  \includegraphics[width=\linewidth]{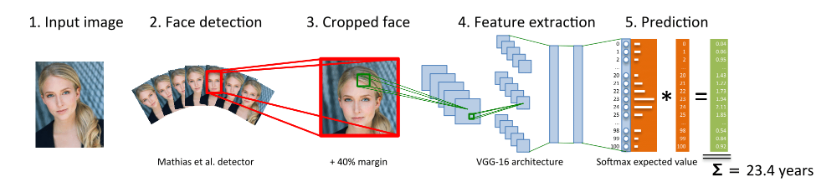}
  \caption{Processing pipeline for age estimation \cite{rothe2016}.}
  \label{fig:pipeline}
\end{figure}

Most age estimation models follow a similar processing pipeline (Fig.~\ref{fig:pipeline}). 
First, a face is detected in the input image, after which the facial region is cropped with an appropriate margin.

The cropped face is then normalized and passed through a convolutional neural network (such as VGG-16 or ResNet) to extract facial features. 
Based on the extracted feature representation, the final age is predicted, typically through a regression or classification layer.

A critical step in this pipeline is the \textbf{cropping and resizing} of the input images, as this determines which visual details are preserved and made available to the feature extractor. 
The effectiveness of this step depends strongly on the resolution and quality of the original input image.
\subsection{Datasets for Age Estimation}

Large, well-labeled datasets are essential for training and evaluating age estimation models. 
These datasets vary in size, age distribution, diversity, and label quality, all of which directly affect the performance of deep learning models.

\begin{enumerate}
  \item \textbf{IMDB-WIKI} \cite{imdbwiki2015} is the largest publicly available dataset for age and gender prediction, containing more than 500,000 images collected from IMDb and Wikipedia. 
  The dataset spans ages from 0 to over 100 years, but suffers from an imbalanced age distribution, with a strong overrepresentation of young adults (15–35 years). 
  Age labels are derived from the estimated date the photo was taken, making them chronological but also prone to noise due to incorrect metadata.

  \item To reduce this label noise, the cleaned subset \textbf{IMDB-Clean} was introduced \cite{imdbclean2021}. 
  This subset removes incorrect or corrupted samples (e.g., blank images, non-face scenes, or incorrect age labels). 
  As a result, IMDB-Clean provides more consistent annotations and is more suitable for experimental research, including this study.

\begin{figure}[h]
  \centering
  \includegraphics[width=0.48\textwidth]{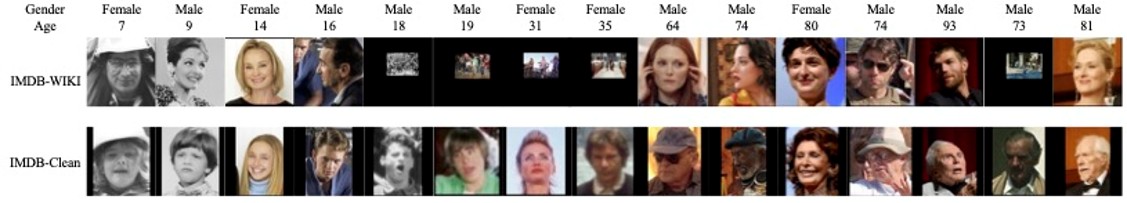}
  \caption{Comparison of sample images from IMDB-WIKI (top) and the cleaned IMDB-Clean dataset (bottom). Incorrect or irrelevant samples have been removed, improving overall data quality.}
  \label{fig:imdb-clean}
\end{figure}

  \item \textbf{UTKFace} \cite{zhang_age_2017} contains approximately 20,000 images with ages ranging from 0 to 116 years, labeled with age, gender, and ethnicity. 
  Due to its wide age range and relatively balanced distribution, UTKFace is frequently used as a benchmark. 
  However, its image resolution (200×200 px) is relatively low.

  \item \textbf{FairFace} \cite{karkkainen2021fairface} focuses specifically on reducing demographic bias. 
  The dataset includes more than 100,000 images and is balanced across ethnicities and genders. 
  However, age labels are provided in broad categories (e.g., 0–2, 3–9, 10–19, etc.) rather than exact ages, which limits its usefulness for regression-based age estimation.

  \item \textbf{VGGFace2} \cite{cao2018vggface2} consists of 3.31 million high-resolution images of 9,131 identities, 
  with considerable variation in pose, age, illumination, and ethnicity. 
  Although the dataset does not provide explicit age labels, it is widely used to train robust face feature extractors (e.g., ResNet and SENet), 
  which are later applied to tasks including age estimation.
\end{enumerate}

\begin{table}[h]
\centering
\caption{Overview of commonly used datasets for age estimation}
\label{tab:datasets}
\renewcommand{\arraystretch}{1.2}

\resizebox{\linewidth}{!}{%
\begin{tabular}{|l|c|p{3.5cm}|p{2.5cm}|}
\hline
\textbf{Dataset} & \textbf{Number of Images} & \textbf{Resolution / Quality} & \textbf{Age Range} \\ \hline
UTKFace         & 20,000+ & $\sim 200 \times 200$ px (low)            & 0--116 \\ \hline
FairFace        & 100,000+ & $\sim 300 \times 300$ px (medium)        & Categories (0–2 to 70+) \\ \hline
IMDB-WIKI       & 500,000+ & often $>$400 px (mixed)                  & 0--100+ \\ \hline
IMDB-Clean      & Subset of IMDB-WIKI & up to 1024 px, cleaned (high) & 0--100+ \\ \hline
VGGFace2        & 3.31M   & high-resolution web images (mixed) & No age labels \\ \hline
\end{tabular}
}
\end{table}

\textbf{In summary:} IMDB-WIKI and UTKFace are among the most widely used datasets for age estimation. 
However, because IMDB-WIKI contains substantial label noise, this study uses the cleaned \textbf{IMDB-Clean} subset \cite{imdbclean2021}. 
A key advantage of IMDB-Clean is that it provides higher-resolution images (up to 1024 px) with more reliable age annotations. 
Although UTKFace is widely used and offers a balanced age distribution, the dataset consists primarily of low-resolution images (200×200 px). 
Therefore, UTKFace is less suitable for analyzing the impact of input resolution, as it does not provide sufficiently high-resolution samples for comparison. 
For this reason, IMDB-Clean is the most appropriate dataset for the objectives of this study.

\subsection{Choice of Existing Frameworks}

For this study, \textit{DeepFace} \cite{serengil_benchmark_2024} and \textit{InsightFace} \cite{insightface2024} were selected because both frameworks provide pretrained age estimation models that can be used directly without additional training. 
Each framework integrates detection, alignment, feature extraction, and regression within a standardized pipeline, enabling reproducible comparisons across different input resolutions.

\textbf{DeepFace} is a Python framework for face recognition and related analysis tasks, including age and gender prediction. 
It supports several state-of-the-art backbone models, such as VGG-Face, ArcFace, FaceNet, Dlib, and DeepID~\cite{serengil_benchmark_2024}. 
The standard pipeline consists of face detection, alignment, normalization, and embedding extraction. 
For age estimation, cropped faces are resized (typically to $224 \times 224$ pixels for VGG-Face or $112 \times 112$ pixels for ArcFace) and passed to a regression model that outputs a predicted age~\cite{parkhi2015}. 
According to the official documentation, DeepFace reports an average error of approximately $\pm 4.65$ years (MAE) on public test data~\cite{deepfacegithub}.

\begin{figure}[h]
    \centering
    \includegraphics[width=0.45\textwidth]{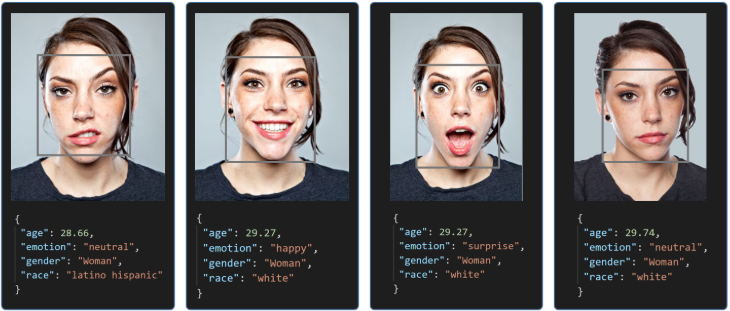}
    \caption{Example output from DeepFace: age, emotion, gender, and race predictions generated from a facial image \cite{serengil_benchmark_2024}.}
    \label{fig:deepface-example}
\end{figure}

\textbf{InsightFace} is an open-source framework originally developed for face recognition and built around the ArcFace architecture~\cite{dengArcFace2019}. 
It provides built-in support for face detection using RetinaFace and SCRFD, both included in the official implementation~\cite{insightface2024}. 
Input images are resized by default to $112 \times 112$ pixels for processing by the ArcFace backbone. 
For age estimation, InsightFace includes dedicated models with a \texttt{genderage} head, in which face embeddings are first extracted and subsequently passed through a regression or classification layer to estimate age and gender.

\textbf{Alternative Approaches Considered:} 
Alternative approaches such as training custom CNN models from scratch or using OpenCV- or YOLO-based age estimation systems were considered but not selected. These alternatives generally require extensive training data and long training times, which fall outside the scope of this project. 

\textbf{Framework Selection:} DeepFace and InsightFace were selected because 
they provide well-documented, pretrained age estimation models with integrated 
pipelines for detection, alignment, and inference. These properties enable 
consistent and reproducible evaluation across different input resolutions, 
making both frameworks suitable for the objectives of this study.

\begin{table}[h]
\centering
\caption{Comparison of DeepFace and InsightFace (2025)}
\label{tab:frameworks}
\renewcommand{\arraystretch}{1.2}
\resizebox{\linewidth}{!}{%
\begin{tabular}{|l|p{4cm}|p{4cm}|}
\hline
\textbf{Property} & \textbf{DeepFace}~\cite{serengil_benchmark_2024} & \textbf{InsightFace}~\cite{insightface2024} \\ \hline
Open source         & MIT License & MIT License \\ \hline
Backbones           & VGG-Face, Facenet, OpenFace, DeepID, ArcFace, Dlib & ArcFace, Partial FC, SubCenter ArcFace \\ \hline
Detection models    & OpenCV, SSD, Dlib, MTCNN, RetinaFace & RetinaFace, SCRFD \\ \hline
Input format        & $224\times224$ (VGG-Face), $112\times112$ (ArcFace) & $112\times112$ (ArcFace) \\ \hline
Updates             & Active (2025, v0.0.95) & Active (2025, v0.7.3) \\ \hline
\end{tabular}
}
\end{table}

\subsection{Architectures for Age Estimation}

Many age estimation models rely on convolutional neural networks (CNNs). 
Two commonly used backbone architectures in existing frameworks are \textbf{VGG-Face} and \textbf{ArcFace}.

The \textbf{VGG-Face} model~\cite{parkhi2015} is based on the VGG-16 architecture and uses input images of 
$224 \times 224$ pixels. The network contains approximately 134 million parameters and produces 4096-dimensional embeddings~\cite{serengil_benchmark_2024}. 
In frameworks such as DeepFace, VGG-Face is repurposed as a feature extractor, with an additional regression layer used for age prediction~\cite{qawaqneh2017}. 
VGG-Face represents a classical, sequential CNN without residual connections and is computationally expensive.

\textbf{ArcFace}~\cite{dengArcFace2019}, on the other hand, is based on a ResNet architecture with residual connections 
and uses input images of $112 \times 112$ pixels. It contains roughly 34 million parameters, generates 512-dimensional embeddings, 
and introduces the \textit{Additive Angular Margin Loss} to produce more discriminative face representations. 
Within InsightFace, ArcFace is extended with a \texttt{genderage} head, enabling direct prediction of age and gender. 
In summary, VGG-Face and ArcFace represent two different generations of CNN architectures: a classical, heavy CNN (VGG-Face) versus a more modern and efficient ResNet-based model (ArcFace). 
This architectural difference partially explains the superior performance and higher speed of InsightFace observed in the experiments of this study. 
Although more recent architectures exist such as FusionNet~\cite{wang2018fusionnet} or attention-based models~\cite{wang2021adpf} these fall outside the scope of this work.

\begin{table}[h]
\centering
\caption{Architectures of commonly used face recognition models \cite{serengil_benchmark_2024}}
\label{tab:architectures}

\renewcommand{\arraystretch}{1.2}
\resizebox{\linewidth}{!}{%
\begin{tabular}{|l|c|c|c|c|}
\hline
\textbf{Model} & \textbf{Input Shape} & \textbf{Embedding Dim.} & \textbf{Parameters} & \textbf{Number of Layers} \\ \hline
FaceNet-128d   & 160 $\times$ 160 $\times$ 3 & 128  &  $\sim$22M   & 447 \\ \hline
VGG-Face       & 224 $\times$ 224 $\times$ 3 & 4096 &  $\sim$134M  & 36  \\ \hline
ArcFace        & 112 $\times$ 112 $\times$ 3 & 512  &  $\sim$34M   & 162 \\ \hline
Dlib           & 150 $\times$ 150 $\times$ 3 & 128  &  $\sim$63M   & 34  \\ \hline
OpenFace       & 96 $\times$ 96 $\times$ 3   & 128  &  $\sim$3M    & 166 \\ \hline
\end{tabular}%
}
\end{table}

\textbf{Impact of Image Resolution}:  
Factors such as resolution and image sharpness play a crucial role in age estimation, as subtle details (e.g., wrinkles and skin texture) can be lost at lower resolutions. 
The study \textit{“Effects of Varying Resolution on Performance of CNN based Image Classification”} \cite{kannojia2018} shows that image resolution has a significant influence on CNN performance. 
Accuracy decreases notably when a model is trained on high-resolution images but tested on lower-resolution data.

\begin{table}[ht]
\centering
\renewcommand{\arraystretch}{1.3}
\caption{Effect of varying image resolution on CNN performance on CIFAR-10. 
Accuracy, precision, and F1-score decrease significantly at lower resolutions~\cite{kannojia2018}.}
\resizebox{\linewidth}{!}{%
\begin{tabular}{|c|c|c|c|c|c|c|}
\hline
\multirow{2}{*}{\textbf{CIFAR-10 Dataset Resolution}} & 
\multicolumn{3}{c|}{\textbf{TOTV}} & 
\multicolumn{3}{c|}{\textbf{TVTV}} \\ \cline{2-7}
 & \textbf{Accuracy} & \textbf{Precision} & \textbf{F1 Score} 
 & \textbf{Accuracy} & \textbf{Precision} & \textbf{F1 Score} \\ \hline
28$\times$28 & 0.9927 & 0.99269 & 0.99263 & 0.9927 & 0.99269 & 0.99263 \\ \hline
21$\times$21 & 0.9905 & 0.99062 & 0.99045 & 0.9924 & 0.99247 & 0.99234 \\ \hline
14$\times$14 & 0.9773 & 0.97828 & 0.97749 & 0.9854 & 0.98586 & 0.98544 \\ \hline
7$\times$7   & 0.6791 & 0.79071 & 0.68767 & 0.7770 & 0.84433 & 0.78416 \\ \hline
\end{tabular}%
}
\label{tab:resolution-impact}
\end{table}

Table~\ref{tab:resolution-impact} confirms that image resolution is a critical factor when using pretrained frameworks for age estimation such as \textit{DeepFace} \cite{serengil_benchmark_2024} and \textit{InsightFace} \cite{insightface2024}. 
Similar effects have been observed in other domains; for example, in medical image analysis (e.g., endoscopy), lower image resolution also leads to a clear drop in CNN performance~\cite{thambawita2021}. 
For this reason, the present study explicitly investigates how different input image resolutions affect the accuracy of age predictions in DeepFace and InsightFace.

\subsection{Dataset and Preprocessing}

For this study, a customized subset of the IMDB-Clean dataset\footnote{\url{https://www.kaggle.com/datasets/yuulind/imdb-clean}} was used.  
The selection consists of 1000 images that meet the following criteria:
\begin{itemize}
    \item captured in 2010 or later (based on file metadata),
    \item at least 1024 pixels in width or height,
    \item containing exactly one face (detected using the default InsightFace model: {\scriptsize\texttt{FaceAnalysis(allowed\_modules=["detection"])}}).
\end{itemize}

Filtering was performed using a Python script that automatically removed images with no detectable face or with multiple faces.  
The final dataset is publicly available on Hugging Face.\footnote{\url{https://huggingface.co/datasets/codershiyar/imdb-clean-2010plus-singleface}}

Each original image was resized into seven versions with the following resolutions:  
$64\times64$, $112\times112$, $224\times224$, $256\times256$, $512\times512$, $720\times720$, and $1080\times1080$ pixels.  
The \texttt{Pillow} library was used for resizing. Since all images are square, no letterboxing (padding) was required.

All versions of each image were stored in a single directory, with the resolution appended to the filename (e.g., \texttt{\_64.jpg}, \texttt{\_512.jpg}).

\begin{figure}[h]
  \centering
  \includegraphics[width=\linewidth]{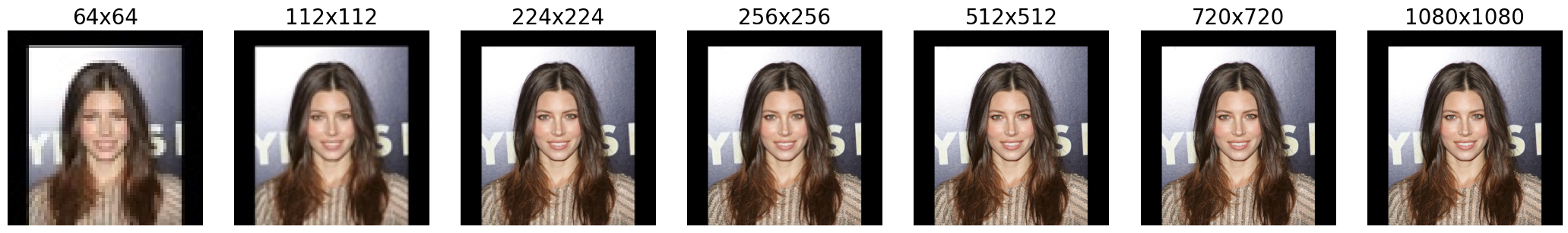}
  \caption{Example of a single image in seven different resolutions.}
  \label{fig:resolutie-voorbeeld}
\end{figure}

\subsection{Experimental Environment}

All experiments were conducted in Google Colab using an NVIDIA T4 GPU and Python 3 as the runtime environment (see Fig.~\ref{fig:colab_gpu}).  
The frameworks DeepFace (v0.0.95) and InsightFace (v0.7.3) were installed via \texttt{pip}:

\begin{lstlisting}[basicstyle=\scriptsize\ttfamily]
!pip install deepface==0.0.95 insightface==0.7.3 
!pip install onnxruntime-gpu-1.23.0
\end{lstlisting}

\begin{figure}[h]
    \centering
    \includegraphics[width=0.3\textwidth]{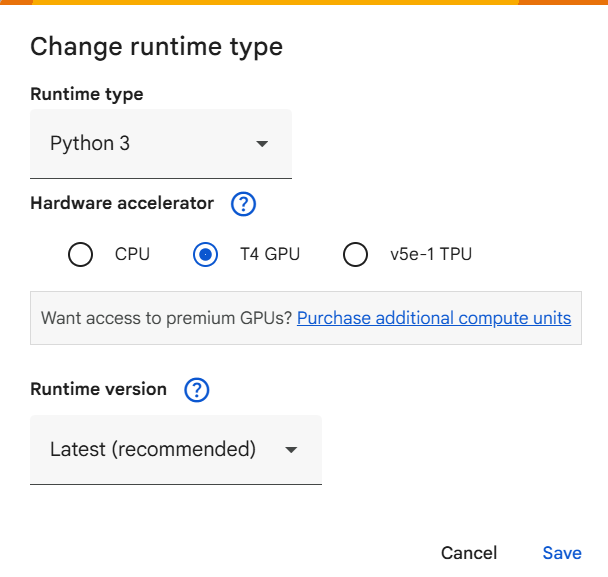}
    \caption{Screenshot of the Colab runtime (T4 GPU, Python 3).}
    \label{fig:colab_gpu}
\end{figure}

For DeepFace, age prediction was performed using:
\begin{lstlisting}[language=Python,basicstyle=\scriptsize\ttfamily]
DeepFace.analyze(img_path=img_path, actions=["age"], 
                 enforce_detection=False)
\end{lstlisting}
This uses the default DeepFace backbone for age estimation, namely VGG-Face, as implemented in version 0.0.95.

For InsightFace, the following configuration was used:
\begin{lstlisting}[language=Python,basicstyle=\scriptsize\ttfamily]
FaceAnalysis(allowed_modules=["detection", "genderage"])
\end{lstlisting}

In this setup, face detection is performed using SCRFD, and age estimation is handled by the built-in \texttt{genderage} head operating on the ArcFace backbone.  
The full code and processing pipeline are publicly available~\cite{coderepository}.

\subsection{Experimental Design}

All experiments used the same 1000 images (the selected subset of IMDB-Clean).  
For each image, seven different resolutions were evaluated ($64 \times 64$ up to $1080 \times 1080$).  
Each resolution was tested using both frameworks:
\begin{itemize}
    \item \textbf{DeepFace (v0.0.95)} – default detection and pretrained regression-based age analysis.
    \item \textbf{InsightFace (v0.7.3)} – default detection (SCRFD) and age estimation via the built-in \texttt{genderage} head.
\end{itemize}

Applying both frameworks systematically across all resolutions ensures a fair comparison.  
Performance was evaluated using MAE, SD, and MedAE (see Section~\ref{sec:evaluatiemetrics}).
\subsection{Evaluation Metrics}
\label{sec:evaluatiemetrics}

To assess the performance of DeepFace and InsightFace across different resolutions, the \textbf{Mean Absolute Error (MAE)} was computed for each model–resolution combination. 
MAE represents the average difference, in years, between the predicted and true age.

In addition, two supplementary statistics were included:

\begin{itemize}
    \item \textbf{Standard Deviation (SD)} – measures the variability of the absolute errors. 
    A high SD indicates irregular or unstable predictions, while a low SD indicates consistent performance.

    \item \textbf{Median Absolute Error (MedAE)} – the median absolute error. 
    This metric is more robust than the mean and highlights whether MAE is influenced by outliers.
\end{itemize}

Together, MAE, SD, and MedAE provide a more complete picture of both the accuracy and the reliability of the age predictions at each resolution.

All calculations were performed on the exact same set of 1000 faces per resolution, ensuring a fair comparison across models and resolutions.  
The final results are summarized in Table~\ref{tab:accuracy_metrics}.

\subsection{Formulas Used}

For each resolution and each model, the following formulas were applied, where $y_i$ denotes the predicted age, $\hat{y}_i$ the ground-truth age, and $n$ the number of images:

\[
MAE = \frac{1}{n} \sum_{i=1}^{n} \left| y_i - \hat{y}_i \right|
\]

\[
SD = \sqrt{\frac{1}{n} \sum_{i=1}^{n} \left(\left| y_i - \hat{y}_i \right| - MAE\right)^2}
\]

\[
MedAE = \mathrm{median} \left(\left| y_1 - \hat{y}_1 \right|, \dots, \left| y_n - \hat{y}_n \right|\right)
\]

\subsection{Processing Time Measurement}
For both frameworks (DeepFace and InsightFace), 100 images were analyzed per resolution (from $64 \times 64$ to $1080 \times 1080$ pixels), resulting in a total of 700 images per framework. 
Each image was processed ten times to minimize variation caused by background processes or system activity. 
For each image, the mean processing time and standard deviation were computed; subsequently, the average of these values was calculated per resolution in seconds (see Table~\ref{tab:runtime}).  
The processing time includes all internal steps performed by the framework, such as face detection and inference, which were executed automatically by the framework.
\section{Results and Discussion}

\begin{figure}[h]
  \centering
  \includegraphics[width=\linewidth]{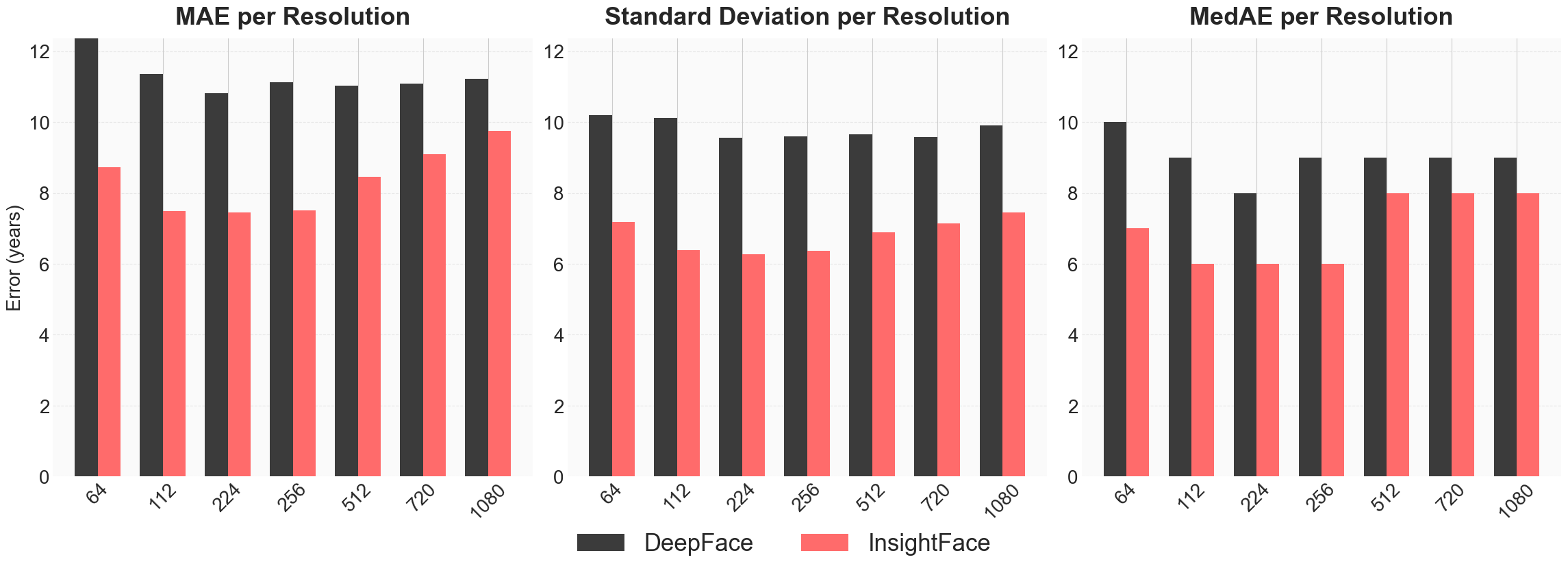}
  \caption{Comparison of DeepFace and InsightFace across different image resolutions based on MAE, SD, and MedAE.}
  \label{fig:resultaten}
\end{figure}

\begin{table}[h]
\centering
\caption{Mean Absolute Error (MAE), Standard Deviation (SD), and Median Absolute Error (MedAE) per resolution.}
\label{tab:accuracy_metrics}
\renewcommand{\arraystretch}{1.1}
\resizebox{\linewidth}{!}{%
\begin{tabular}{|c|ccc|ccc|}
\hline
\multirow{2}{*}{\textbf{Resolution}} & \multicolumn{3}{c|}{\textbf{DeepFace}} & \multicolumn{3}{c|}{\textbf{InsightFace}} \\
 & MAE ↓ & SD ↓ & MedAE ↓ & MAE ↓ & SD ↓ & MedAE ↓ \\ \hline

$64 \times 64$   & 12.36 & 10.19 & 10.0 & 8.72 & 7.19 & 7.0 \\ \hline
$112 \times 112$ & 11.36 & 10.12 & 9.0  & 7.48 & 6.38 & 6.0 \\ \hline
$224 \times 224$ & 10.83 & 9.57  & 8.0  & 7.46 & 6.28 & 6.0 \\ \hline
$256 \times 256$ & 11.13 & 9.60  & 9.0  & 7.51 & 6.36 & 6.0 \\ \hline
$512 \times 512$ & 11.02 & 9.65  & 9.0  & 8.45 & 6.89 & 8.0 \\ \hline
$720 \times 720$ & 11.08 & 9.58  & 9.0  & 9.09 & 7.14 & 8.0 \\ \hline
$1080 \times 1080$ & 11.22 & 9.90 & 9.0 & 9.76 & 7.45 & 8.0 \\ \hline

\end{tabular}%
}
\end{table}

Table~\ref{tab:accuracy_metrics} and Fig.~\ref{fig:resultaten} show the error metrics
(MAE, SD, and MedAE) for each resolution using DeepFace and InsightFace. 
Both frameworks achieve their lowest MAE at $224 \times 224$ pixels:  
10.83 years for DeepFace and 7.46 years for InsightFace.

At $64 \times 64$, the error increases to 12.36 (DeepFace, +14.1\%) and 8.72 (InsightFace, +16.9\%) 
compared to $224 \times 224$. Previous work suggests that this may be related to the loss of 
age-relevant facial details such as wrinkles and skin texture when the resolution is too low~\cite{kannojia2018}.

At very high resolutions ($1080 \times 1080$), DeepFace remains relatively stable (11.22 years; +3.6\%), 
while InsightFace degrades more strongly, reaching 9.76 years (+30.8\%).  
DeepFace remains consistently around $\sim$11 years MAE, whereas InsightFace exhibits a gradual decline in performance when moving above $224 \times 224$.  
This may indicate that InsightFace's internal resizing step handles extreme downscaling from high-resolution images to $112 \times 112$ pixels less effectively, causing loss of useful details.

A notable observation is that InsightFace performs slightly better at an external input resolution of 
$224 \times 224$ (MAE = 7.46 years) than at its native $112 \times 112$ input (MAE = 7.48 years), 
even though the internal resize still converts all inputs to $112 \times 112$.  
This suggests that higher input resolutions may improve face detection (SCRFD) and alignment 
before the internal resizing occurs, thereby improving overall accuracy.

\begin{table}[h]
\centering
\caption{Average error metrics (MAE, SD, MedAE) across all resolutions.}
\label{tab:avg_metrics}
\renewcommand{\arraystretch}{1.2}
\begin{tabular}{|l|c|c|c|}
\hline
\textbf{Framework} & \textbf{MAE ↓} & \textbf{SD ↓} & \textbf{MedAE ↓} \\ \hline
DeepFace    & 11.28 & 9.80 & 9.00 \\ \hline
InsightFace & 8.35  & 6.81 & 7.00 \\ \hline
\end{tabular}
\end{table}

Across all resolutions, InsightFace outperforms DeepFace on average (Table~\ref{tab:avg_metrics}). 
The mean MAE is 8.35 years for InsightFace and 11.28 years for DeepFace. 
In addition to MAE, the Standard Deviation (SD) and Median Absolute Error (MedAE) provide further insight into model behavior. 
For InsightFace at $224 \times 224$ pixels, the SD is 6.28 years and the MedAE is 6.0 years, 
indicating reasonably consistent errors and showing that half of the predictions deviate by at most 6 years from the true age. 
DeepFace exhibits higher SD and MedAE values, suggesting greater variability and lower typical accuracy.

These differences can be linked to the underlying architectures:  
InsightFace uses ArcFace with a ResNet backbone and Additive Angular Margin Loss,  
whereas DeepFace relies on VGG-Face with softmax loss~\cite{dengArcFace2019,parkhi2015}.

\textbf{Processing time:}  
Although the primary research question focuses on accuracy, processing time was also examined to assess the computational efficiency of both frameworks at different resolutions. 
This is relevant for real-time applications and systems with limited computational resources. 
Table~\ref{tab:runtime} reports the average per-image processing time for both frameworks across all resolutions.

InsightFace is substantially faster and remains nearly constant between 0.015 s and 0.021 s per image across all resolutions.  
DeepFace, however, shows a clear increase in processing time as resolution increases from an average of 0.052 ± 0.070 s at $64\times64$ to 0.733 ± 0.106 s at $1080\times1080$.  
Notably, DeepFace is slightly slower at $64\times64$ than at $112\times112$, which may be due to internal upscaling requirements or less stable detection at very low resolutions.

Considering the combination of lower error rates (Table~\ref{tab:accuracy_metrics}) and significantly higher processing speed,  
InsightFace can be considered both \textbf{more accurate} and \textbf{more efficient} than DeepFace in these experiments.

\begin{table}[h]
\centering
\caption{Average processing time per resolution (in seconds) for DeepFace and InsightFace, including standard deviation.}
\label{tab:runtime}
\renewcommand{\arraystretch}{1.1}
\begin{tabular}{|c|c|c|}
\hline
\textbf{Resolution} & \textbf{DeepFace (Mean ± SD)} & \textbf{InsightFace (Mean ± SD)} \\ \hline
64×64     & 0.052 ± 0.070 s & 0.016 ± 0.006 s \\ \hline
112×112   & 0.039 ± 0.002 s & 0.015 ± 0.001 s \\ \hline
224×224   & 0.074 ± 0.005 s & 0.016 ± 0.001 s \\ \hline
256×256   & 0.086 ± 0.007 s & 0.016 ± 0.001 s \\ \hline
512×512   & 0.222 ± 0.022 s & 0.017 ± 0.001 s \\ \hline
720×720   & 0.381 ± 0.044 s & 0.019 ± 0.001 s \\ \hline
1080×1080 & 0.733 ± 0.106 s & 0.021 ± 0.001 s \\ \hline
\end{tabular}
\end{table}

\section{Conclusion}

This study demonstrates that input image resolution has a clear and measurable impact on the accuracy of age estimation using DeepFace and InsightFace, despite their internal preprocessing steps. 
Both frameworks achieve optimal performance at an input resolution of $224 \times 224$ pixels, yielding the lowest Mean Absolute Error (MAE). 
Deviations from this resolution, both lower and higher, lead to reduced accuracy. InsightFace is more sensitive to extremely high resolutions (+31\% at $1080 \times 1080$) than DeepFace (+3.6\%).

A notable finding is that InsightFace performs slightly better with $224 \times 224$ pixel inputs (MAE: 7.46 years) compared to its native $112 \times 112$ input (MAE: 7.48 years), even though all images are internally resized to $112 \times 112$. 
This indicates that the external input resolution still influences the final prediction quality.

In addition to achieving higher accuracy, InsightFace also demonstrates substantially faster processing times across all resolutions, reflecting a more efficient underlying architecture. 
Because the IMDB-Clean dataset consists of images of celebrities, the demographic distribution (e.g., age, gender, ethnicity) may not fully represent the general population, which could limit the generalizability of the findings.

Future research could therefore focus on evaluating DeepFace and InsightFace across different demographic groups to identify potential bias in age estimation. 
Considering the balance between speed and accuracy, InsightFace with an input resolution of $224 \times 224$ pixels is recommended for real-time applications and systems with limited computational resources.

\end{document}